\documentclass[10pt,twocolumn,letterpaper]{article}

\usepackage{cvpr}              
\usepackage[accsupp]{axessibility}

\usepackage{graphicx}
\usepackage{amsmath}
\usepackage{amssymb}
\usepackage{booktabs}
\usepackage{enumitem}
\newcommand{\system}{ObjectFormer\xspace}
\usepackage{multicol}
\usepackage[ruled]{algorithm2e}

\usepackage{multirow}
\usepackage{amsthm}
\usepackage{mathrsfs}
\usepackage{diagbox}
\usepackage{bm}
\usepackage{newfloat}
\usepackage{xspace}
\usepackage{xcolor}

\newcommand{\Drop}[1]{\textcolor{red}{\xspace\small{\bf $\downarrow$#1}}}

\makeatletter
\newcommand\figcaption{\def\@captype{figure}\caption}
\newcommand\tabcaption{\def\@captype{table}\caption}
\makeatother

\definecolor{green_im}{rgb}{0.1, 0.55, 0.3}

\definecolor{citecolor}{RGB}{0, 113, 188}

\usepackage[pagebackref,breaklinks,colorlinks, citecolor=citecolor]{hyperref}

\usepackage[capitalize]{cleveref}
\crefname{section}{Sec.}{Secs.}
\Crefname{section}{Section}{Sections}
\Crefname{table}{Table}{Tables}
\crefname{table}{Tab.}{Tabs.}

\def\cvprPaperID{1872} 
\def\confName{CVPR}
\def\confYear{2022}

\makeatletter
\def\thanks#1{\protected@xdef\@thanks{\@thanks
        \protect\footnotetext{#1}}}
\makeatother

\begin{document}

\title{\system for Image Manipulation Detection and Localization}

\author{ Junke Wang\textsuperscript{\rm 1,2}, Zuxuan Wu\textsuperscript{\rm 1,2 \dag}, Jingjing Chen\textsuperscript{\rm 1,2}, Xintong Han\textsuperscript{\rm 3}, \\ Abhinav Shrivastava\textsuperscript{\rm 4}, Ser-Nam Lim\textsuperscript{\rm 5},  Yu-Gang Jiang\textsuperscript{\rm 1,2} \\
\textsuperscript{\rm 1}Shanghai Key Lab of Intelligent Information Processing, School of Computer Science, Fudan University \\ 
\textsuperscript{\rm 2}Shanghai Collaborative Innovation Center on Intelligent Visual Computing \\ 
\textsuperscript{\rm 3}Huya Inc,
 \textsuperscript{\rm 4}University of Maryland, \textsuperscript{\rm 5}Meta AI 
 \thanks{\dag Corresponding author.}
}

\maketitle

\begin{abstract}
Recent advances in image editing techniques have posed serious challenges to the trustworthiness of multimedia data, which drives the research of image tampering detection. In this paper, we propose \system to detect and localize image manipulations. To capture subtle manipulation traces that are no longer visible in the RGB domain, we extract high-frequency features of the images and combine them with RGB features as multimodal patch embeddings. Additionally, we use a set of learnable object prototypes as mid-level representations to model the object-level consistencies among different regions, which are further used to refine patch embeddings to capture the patch-level consistencies. We conduct extensive experiments on various datasets and the results verify the effectiveness of the proposed method, outperforming state-of-the-art tampering detection and localization methods.
\end{abstract}


\section{Introduction}
\label{sec:intro}
With the rapid development of deep generative models like GANs~\cite{goodfellow2014generative, mirza2014conditional, zhu2017unpaired} and VAEs~\cite{kingma2013auto, razavi2019generating}, a multitude of image editing applications have become widely accessible to the public~\cite{park2020swapping, vinker2020deep, dhamo2020semantic, li2020manigan}. These editing tools make it easy and effective to produce photo-realistic images and videos that could be used for entertainment, interactive design, \etc,  which otherwise requires professional skills. However, there are growing concerns on the abuse of editing techniques to manipulate image and video content for malicious purposes. Therefore, it is crucial to develop effective image manipulation detection methods to examine whether images have been modified or not and identify regions in images that have been modified.

\begin{figure}[t]
  \centering
   \includegraphics[width=1.0\linewidth]{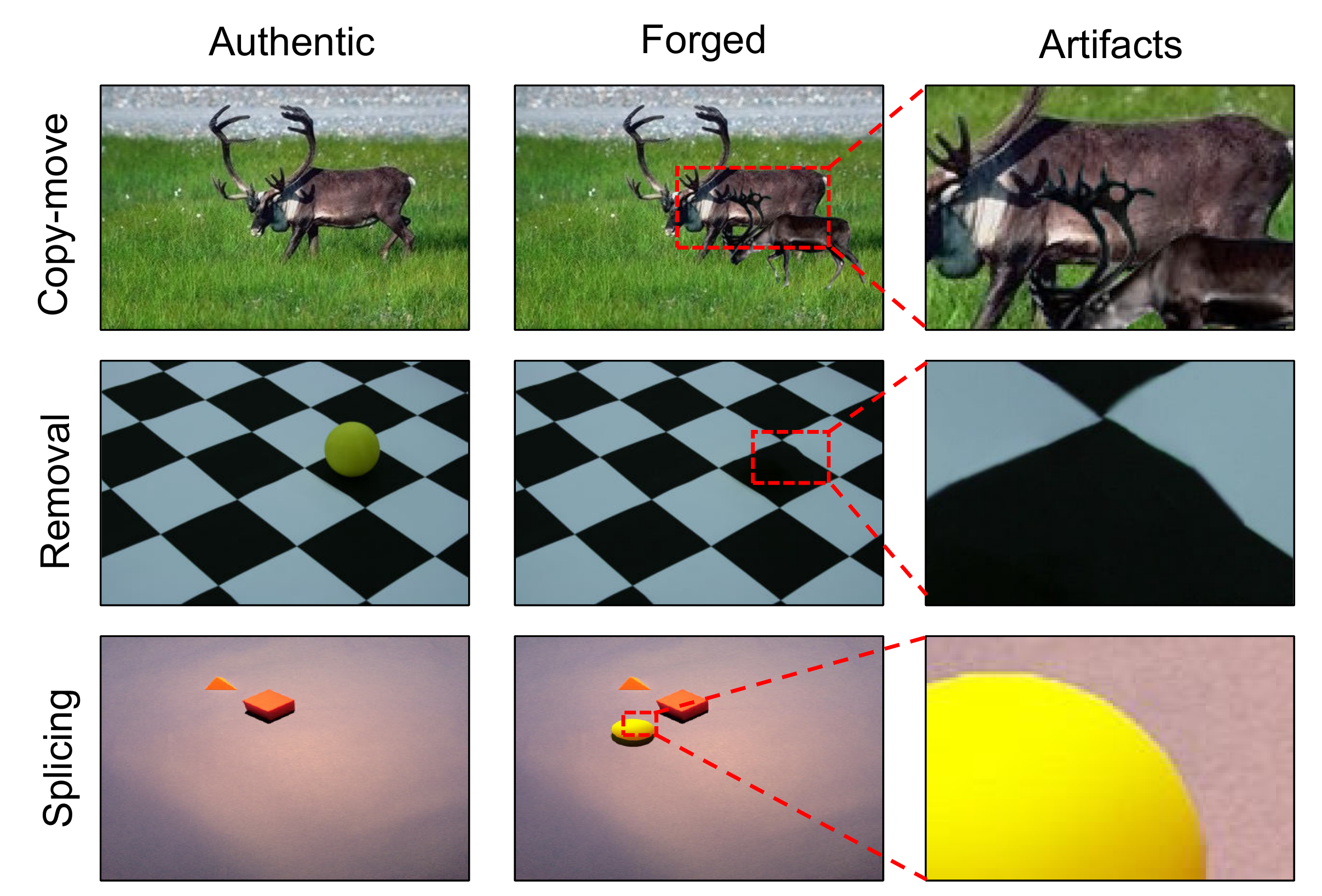}
   \vspace{-0.2in}
   \caption{Tampering images usually contain manipulated objects. Thus, exploiting object-level consistency is crucial for manipulation detection.}
   \vspace{-0.2in}
   \label{fig:example}
\end{figure}

Image manipulation techniques can be generally classified into three types: (1) splicing methods which copy regions from one image and paste to other images, (2) copy-move which shifts the spatial locations of objects within images, and (3) removal methods which erase regions from images and inpaint missing regions with visually plausible contents. As shown in Figure~\ref{fig:example}, to produce semantically meaningful and perceptually convincing images, these approaches oftentimes manipulate images at the object-level, \ie, adding/removing objects in images.  While there are some recent studies focusing on image manipulation detection~\cite{zhou2018learning, wu2019mantra, hu2020span}, they typically use CNNs to directly map input images to binary labels (\ie, authentic/manipulated) without explicitly modeling object-level representations. In contrast, we posit that \textbf{image manipulation detection should not only examine whether certain pixels are out of distribution, but also consider whether objects are consistent with each other}.  In addition, visual artifacts brought by image editing that are no longer perceptible in the RGB domain are oftentimes noticeable in the frequency domain~\cite{chen2021local, qian2020thinking, wang2021m2tr}. This demands a multimodal approach that jointly models the RGB domain and the frequency domain to discover subtle manipulation traces.

In this paper, we introduce \system, a multimodal transformer framework for image manipulation detection and localization. \system builds upon transformers due to their impressive performance on a variety of vision tasks like image classification~\cite{dosovitskiy2020image, heo2021rethinking,meng2021adavit}, object detection~\cite{carion2020end, zhu2020deformable}, video classification~\cite{liu2021video, bertasius2021space, wang2021efficient,wang2021bevt}, \etc. More importantly, transformers are natural choices to model whether patches/pixels are consistent in images, given that they explore the correlations between different spatial locations using self-attention. Inspired by object queries that are automatically learned~\cite{zhu2020deformable,bai2021visual}, we use a set of learnable parameters as object prototypes (serving as mid-level object representations) to discover the object-level consistencies, which are further leveraged to refine the patch embeddings for patch-level consistencies modeling.

With this in mind, \system first converts an image from the RGB domain to the frequency domain using Discrete Cosine Transform and then extracts multimodal patch embeddings with a few convolutional layers. 
The RGB patch embeddings and the frequency patch embeddings are further concatenated to complement each other. Further, we use a set of learnable embeddings as object queries/prototypes, interacting with the derived patch embeddings to learn consistencies among different objects. We refine patch embeddings with these object prototypes with cross-attention. By iteratively doing so, \system derives global feature representations that explicitly encode mid-level object features, which can be readily used to detect manipulation artifacts. Finally, the global features are used to predict whether images have been modified and the corresponding manipulation mask in a multi-task fashion. The framework can be trained in an end-to-end manner. We conduct experiments on commonly used image tampering datasets, including CASIA~\cite{dong2013casia}, Columbia~\cite{shi2000normalized}, Coverage~\cite{wen2016coverage}, NIST16~\cite{nimble16}, and IMD20~\cite{novozamsky2020imd2020}. The results demonstrate that \system outperforms state-of-the-art tampering detection and localization methods. In summary, our work makes the following key contributions:
\begin{itemize}
\item We introduce \system, an end-to-end multimodal framework for image manipulation detection and localization, combining RGB features and frequency features to identify the tampering artifacts. 
\item We explicitly leverage learnable object prototypes as mid-level representations to model object-level consistencies and refined patch embeddings to capture patch-level consistencies. 
\item We conduct extensive experiments on multiple benchmarks and demonstrate that our method achieves state-of-the-art detection and localization performance.
\end{itemize}

\begin{figure*}[t]
  \centering
  \includegraphics[width=\linewidth]{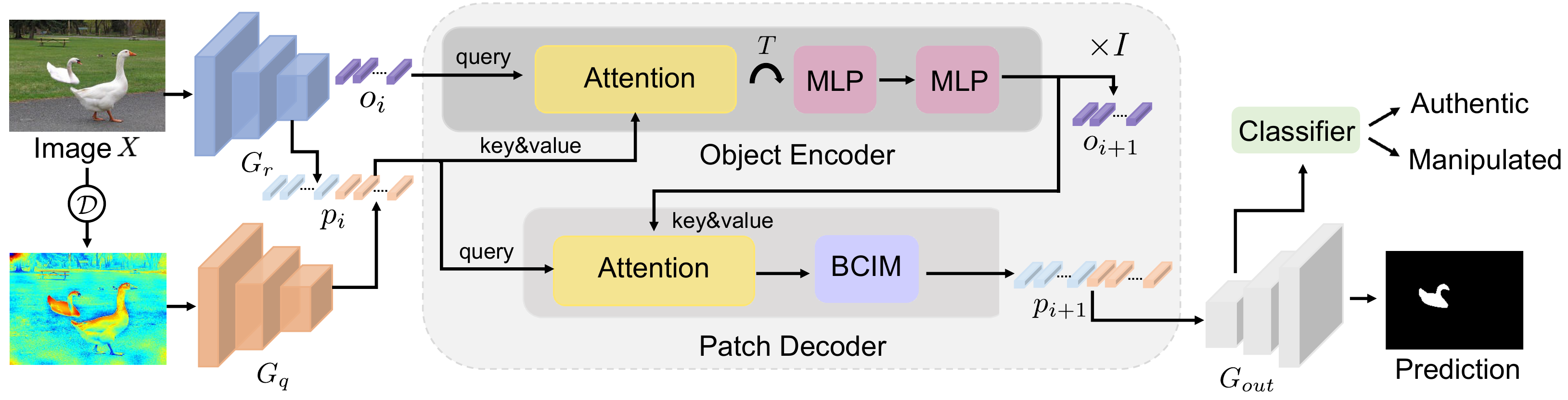}
   \vspace{-0.2in}
  \caption{An overview of \system. The input is a suspicious image ($H \times W \times 3$), and the output includes a tampering localization result and a predicted mask ($H \times W \times 1$), which localizes the manipulation regions.}
  \vspace{-0.1in}
  \label{fig:networks}
\end{figure*}

\section{Related Work}
\label{sec:related}

\noindent \textbf{Image Manipulation Detection / Localization}
Most early studies focus on detecting a specific type of manipulation, \eg, splicing~\cite{cozzolino2015splicebuster, huh2018fighting, kniaz2019point}, copy-move~\cite{cozzolino2015efficient,rao2016deep}, and removal~\cite{zhu2018deep}. However, in real-world scenarios, the exact manipulation type is unknown, which motivates a line of work focusing on general manipulation detection~\cite{wu2019mantra, bappy2019hybrid, hu2020span}. In addition, RGB-N~\cite{zhou2018learning} introduces a two-stream network for manipulation localization, where one stream extracts RGB features to capture visual artifacts, and the other stream leverages noise features to model the inconsistencies between tampered and untouched regions. SPAN~\cite{hu2020span} models the relationships of pixels within image patches on multiple scales through a pyramid structure of local self-attention blocks. PSCCNet~\cite{liu2021pscc} extracts hierarchical features with a top-down path and detects whether input image has been manipulated using a bottom-up path. In this work, we detect the manipulation artifacts by explicitly adopting a set of learnable embeddings as object prototypes for object-level consistency modeling and the refined patch embeddings for patch-level consistency modeling.

\vspace{0.1in}
\noindent \textbf{Visual Transformer}
The remarkable success of Transformers~\cite{vaswani2017attention} and their variants in natural language processing has motivated a plethora of work exploring transformers for a variety of computer vision tasks due to their capabilities in modeling long-range dependencies. More specifically, ViT~\cite{dosovitskiy2020image} reshapes an image into a sequence of flattened patches and inputs them to the transformer encoders for image classification. T2T~\cite{yuan2021tokens} incorporates a Token-to-Token module to progressively aggregate local information before using self-attention layers. There are also some studies combining the self-attention blocks with classical convolutional neural networks. For instance, DETR~\cite{carion2020end} uses the pretrained CNN for extracting low-level features, which are then fed into a transformer-based encoder-decoder architecture for object detection. In contrast, we introduce the frequency information to facilitate the capture of subtle forgery traces, which are further combined with RGB modality features through a multi-modal transformer for image tampering detection.

\section{Method}
\label{sec:method}
Our goal is to detect manipulated objects within images by modeling visual consistencies among mid-level representations, which are automatically derived by attending to multimodal inputs. In this section, we introduce \system, which consists of a High-frequency Feature Extraction Module (Section~\ref{sec:high_freq}), an object encoder (Section~\ref{sec:obj_en}) that uses learnable object queries to learn whether mid-level representations in images are coherent, and a patch decoder (Section~\ref{sec:img_en}) that produces refined global representations for manipulation detection and localization. Figure~\ref{fig:networks} gives an overview of the framework. 

More formally, we denote an input image as \textit{X} $\in$ $\mathbb{R}^{H \times W \times 3}$, where $\textit{H}$ and $\textit{W}$ are the height and width of the image, respectively. We first extract the feature map $\textit{G}_{r}$ $\in \mathbb{R}^{H_{s} \times W_{s} \times C_{s}}$ and generate patch embeddings using a few convolutional layers, parameterized by $g$, for faster convergence as in~\cite{xiao2021early}. 

\vspace{0.05in}
\subsection{High-frequency Feature Extraction}
\label{sec:high_freq}
As manipulated images are generally post-processed to hide tampering artifacts, it is difficult to capture subtle forgery traces in the RGB space. Therefore, we extract features from the frequency domain to provide complementary clues for manipulation detection.

Taking the image \textit{X} as input, \system first transforms it from the RGB domain to the frequency domain using Discrete Cosine Transform (DCT):
\begin{equation}
X_{q} = \mathcal{D}(X),
\end{equation}
where $X_{q} \in \mathbb{R}^{H \times W \times 1}$ is the frequency domain representation and $\mathcal{D}$ denotes DCT. Then we obtain the high-frequency component through a high pass filter, and transform it back to RGB domain to preserve the shift invariance and local consistency of natural images:
\begin{equation}
X_{h} = \mathcal{D}^{-1}(\mathcal{F}(X_{q}, \alpha)),
\end{equation}
where $\mathcal{F}$ denotes the high pass filter and $\alpha$ is the manually-designed threshold which controls the low frequency component to be filtered out. After that, we input $\textit{X}_{h}$ to several convolutional layers to extract frequency features $G_{f}$, the size of which is the same with $\textit{G}_{r}$. 

We then generate spatial patches of the same sizes using $G_{r}$ and $G_{f}$ and further flatten them to a sequence of $C$-d vectors with the length of $L$. We concatenate the two sequences to obtain a multimodal patch vector $p \in \mathbb{R}^{2L \times C}$.  Sinusoidal positional embeddings~\cite{carion2020end} are added to $p$ to provide positional information.

\vspace{0.05in}
\subsection{Object Encoder}
\label{sec:obj_en}
The object encoder aims to learn a group of mid-level representations automatically that attend to specific regions in $G_{r}$/$G_{f}$ and identify whether these regions are consistent with each other. To this end, we use a set of learnable parameters $o \in \mathbb{R}^{N \times C}$ as object prototypes, which are learned to represent objects that may appear in images. $N$ is a manually designed constant value indicating the maximum number of objects, which we empirically set to 16 in this paper.


Specifically, given the object representations $\textit{o}_{i}$ from the $i$-th layer, we first normalize it with Layer Normalization (LN) and use it
as the query of the attention block. The patch embeddings $\textit{p}_{i}$  after the normalization serve as the key and value. Note that we set $\textit{p}_{0} = p, \textit{o}_{0} = o$, respectively. Then we calculate the object-patch affinity matrix $A_{i} \in \mathbb{R}^{N \times L}$ with matrix multiplication and a \texttt{softmax} function: 
\begin{equation}
\label{eq:affinity}
    A_{i} = \texttt{softmax} \left(\frac{o_{i}W_{eq} \cdot (p_{i}W_{ek})^{T}}{\sqrt{C}}\right),
\end{equation}
where $W_{eq}$ and $W_{ek}$ are  learnable parameters of two linear projection layers. After that, we use another linear layer to project $p_{i}$ into value embedding, and further compute its weighted average with $A_{i}$ to obtain the attention matrix. Finally, the object representations are updated through a residual connection with the attention matrix to obtain $\hat{o}_{i} \in \mathbb{R}^{N \times C}$:
\begin{equation}
    \hat{o}_{i} = o_{i} + A_{i} \cdot p_{i}W_{ev} ,
\end{equation}
where $W_{ev}$ is the learnable parameter for the value embedding layer. With this, each object representation can be injected with global contextual information from all locations. Then we further enable the interaction among different objects using a single linear projection:
\begin{equation}
    \widetilde{o}_{i} = \hat{o}_{i} + (\hat{o}_{i}^{T} W_{c})^{T},
\end{equation}
where $W_{c} \in \mathbb{R}^{N \times N}$ is a learnable weight matrix. This essentially learns how different object prototypes interact with one another to discover object-level visual inconsistencies.

Since the number of objects within an image varies, we additionally use linear projection layers and an activation function GELU~\cite{hendrycks2016gaussian} to enhance the object features. This process can be formulated as:
\begin{equation}
    o_{i+1} = \widetilde{o}_{i} + \delta(\widetilde{o}_{i}W_{act_{1}})W_{act_{2}},
\end{equation}
where $W_{act_{1}}$ and $W_{act_{2}}$ are learnable parameters, $\delta$ is the GELU function, $o_{i+1}$ is the updated object representation.

\vspace{0.05in}
\subsection{Patch Decoder}
\label{sec:img_en}

The object encoder allows different objects within the images to interact with each other to model whether mid-level representations are visually coherent and attend to important patches. In addition to this, we use the updated object representations from the object encoder to further refine the patch embeddings. More specifically, we use $p_{i}$ as query, $o_{i+1}$ as key and value, and enhance the patch features following classic attention paradigm. With this, each patch embedding can further absorb useful information from the derived object prototypes.

More specifically, we first adopt Layer Normalization to normalize both $p_{i}$ and $o_{i+1}$, and then feed them into an attention block for patch embeddings refinement. The complete process can be formulated as:
\begin{equation}
    \begin{split}
     \hat{p}_{i} &= p_{i} + \texttt{softmax} \left(\frac{p_{i}W_{dq} \cdot (o_{i+1}W_{dk})^{T}}{\sqrt{C}} \right) \cdot o_{i+1}W_{dv} , \\
     \overline{p}_{i} &= \hat{p}_{i} + \texttt{MLP}(\hat{p}_{i}),
    \end{split}
\end{equation}
where $W_{dq}$, $W_{dk}$, and $W_{dv}$ are the learnable parameters of three embedding layers, and \texttt{MLP} denotes a Multi-Layer Perceptron that has two linear mappings.

After aggregating the mid-level object features into each patch within the images, we further apply a boundary-sensitive contextual incoherence modeling (\textit{BCIM}) module to detect pixel-level inconsistency for fine-grained feature modeling. In particular, we first reshape $\overline{p}_{i} \in \mathbb{R}^{2N \times C}$ to a 2D feature map $\widetilde{P}_{i}$  with the size $\mathbb{R}^{H_{s} \times W_{s} \times 2C_{s}}$. We then calculate the similarity between each pixel and surrounding pixels within a local window:
\begin{equation}
    S_{i_{j}} = \frac{1}{k \times k} \sum_{j \in \kappa}\texttt{Sim}( \widetilde{P}_{i_{j}},\widetilde{P}_{i_{k}}),
\end{equation}
where $\kappa$ denotes a small $k \times k$ window in the feature map $\widetilde{P}_{i}$,  $\widetilde{P}_{i_{j}}$ is the central feature vector of the window, and $\widetilde{P}_{i_{k}}$ is its neighboring feature vector within $\kappa$. The similarity measurement function \texttt{Sim} that we use is cosine similarity. Then we compute the element-wise summation between $S_{i} \in \mathbb{R}^{H_{s} \times W_{s} \times 1}$ and $\widetilde{P}_{i}$ to obtain a boundary-sensitive feature map with the size of $\mathbb{R}^{H_{s} \times W_{s} \times 2C_{s}}$ to obtain a boundary-sensitive feature map, and finally serialize it to patch embeddings $p_{i+1} \in \mathbb{R}^{2N \times C}$.

Note that we use stacked object encoders and image decoders in a sequential order for $I$ times (which we set to 8 in this paper) to alternately update the object representations and patch features. Finally, we obtain $p_{out} \in \mathbb{R}^{2N \times C}$ which contains visual consistency information at both the object-level and the patch-level. After that, we reshape it to a 2D feature map $G_{out}$, which is then used for manipulation detection and localization.

\vspace{0.05in}
\subsection{Loss Functions}
For manipulation detection, we apply global average pooling on $G_{out}$, and calculate the final binary prediction $\hat{y}$ using a fully connected layer. While for manipulation localization, we progressively upsample $G_{out}$  by alternating convolutional layers and linear interpolation operations to obtain a predicted mask $\hat{M}$. Given the ground-truth label $y$ and mask $M$, we train \system with the following objective function:

\begin{align}
    \mathcal{L} = \mathcal{L}_{cls}(y, \hat{y}) + \lambda \mathcal{L}_{seg}(M, \hat{M}),
    \label{eqn:cls}
\end{align}
where both $\mathcal{L}_{cls}$ and $\mathcal{L}_{seg}$ are binary cross-entropy loss, and $\lambda_{seg}$ is a balancing hyperparameter. By default, we set $\lambda_{seg} = 1$.

\section{Experiments}
\label{sec:exp}
We evaluate our models on two closely related tasks: manipulation localization and
detection. In the former task, our goal is to localize the manipulated regions within the images. In the latter task, the goal is to classify images as being manipulated or authentic. Below, we introduce the experimental setup in Sec.~\ref{sec:setup}, and present results in Sec.~\ref{sec:localization} - Sec.~\ref{sec:robust}, and finally 
we perform an ablation study to justify the effectiveness of different components in Sec.~\ref{sec:ablation}, and show the visualization results in Sec.~\ref{sec:vis}.

\subsection{Experimental Settings}
\label{sec:setup}
\noindent \textbf{Synthesized Pre-training Data} 
We synthesize a large-scale image tampering dataset and pre-train our model on it. The synthesized dataset includes three subsets: 1) FakeCOCO, which is built on MS COCO~\cite{lin2014microsoft}. Inspired by~\cite{zhou2018learning}, we use the annotations provided by MS COCO to randomly copy and paste an object within the same image, or slice an object from one image to another. We also apply the Poisson Blending algorithm between source and target images to erase the slicing boundaries.  2) FakeParis, which is built on Paris StreetView~\cite{pathak2016context} dataset. We erase a region from an authentic image, and adopt a state-of-the-art inpainting method Edgeconnnect~\cite{nazeri2019edgeconnect} to restore visual contents within it. 3) Pristine images, \ie, the original images from the above datasets. We randomly add Gaussian noise or apply the JPEG compression algorithm to the generated data to resemble the visual quality of images in realistic scenarios.

\vspace{0.1in}
\noindent \textbf{Testing Datasets} We follow PSCCNet~\cite{liu2021pscc} to evaluate our model on CASIA~\cite{dong2013casia} dataset, Columbia~\cite{shi2000normalized} dataset,
Carvalho~\cite{wen2016coverage}, Nist Nimble 2016 (NIST16) dataset~\cite{nimble16}, and IMD20~\cite{novozamsky2020imd2020} dataset.

\begin{itemize}[leftmargin=*]
\setlength{\itemsep}{5pt}
\item \textbf{CASIA}~\cite{dong2013casia} provides spliced and copy-moved images of various objects. The tampered regions are carefully selected and some post-processing techniques like filtering and blurring are also applied. Ground-truth masks are obtained by binarizing the difference between tampered and original images. 
\item \textbf{Columbia}~\cite{shi2000normalized} dataset focuses on splicing based on uncompressed images. Ground-truth masks are provided.
\item \textbf{Coverage}~\cite{wen2016coverage} dataset contains 100 images generated by copy-move techniques, the ground-truth masks are also available.
\item \textbf{NIST16}~\cite{nimble16} is a challenging dataset which contains all three tampering techniques. The manipulations in this dataset are post-processed to conceal visible traces. They provide ground-truth tampering mask for evaluation.
\item \textbf{IMD20}~\cite{novozamsky2020imd2020}, which collects 35,000 real images captured by different camera models and generates the same number of forged images using a large variety of Inpainting methods. 
\end{itemize}
To fine-tune \system, we use the same training/testing splits as \cite{hu2020span, liu2021pscc} for fair comparisons.

\vspace{0.05in}
\noindent \textbf{Evaluation Metrics} We evaluate the performance of the proposed method on both image manipulation detection task and localization task. For detection results, we use image-level Area Under Curve (AUC) and F1 score as our evaluation metric, while for localization, the pixel-level AUC and F1 score on manipulation masks are adopted. Since binary masks and detection scores are required to compute F1 scores, we adopt the Equal Error Rate (EER) threshold to binarize them.

\vspace{0.05in}
\noindent \textbf{Implementation Details} All images are resized to $256 \times 256$. For our backbone network, we use EfficientNet-b4~\cite{tan2019efficientnet} pretrained on ImageNet~\cite{deng2009imagenet}. We use Adam for optimization with a learning rate of 0.0001. We train the complete model for 90 epochs with a batch size of 24, and the learning rate is decayed by 10 times every 30 epochs.

\begin{table*}[t]
\begin{minipage}[t]{0.55\linewidth}
  \vspace{0pt}
  \renewcommand{\arraystretch}{1.58}
  \setlength{\tabcolsep}{0pt} 
  \begin{tabular*}{\linewidth}{@{\extracolsep{\fill}}lcccccccc@{}}
    \toprule
    \textbf{Method}  && \textbf{\#Data} && \textbf{Colombia}  & \textbf{Coverage} &  \textbf{CASIA} & \textbf{NIST16} & \textbf{IMD20} \\
    \cmidrule{1-1} \cmidrule{3-3} \cmidrule{5-9}
    ManTraNet && 64K &&  82.4 & 81.9 & 81.7 & 79.5 & 74.8 \\
    SPAN && 96K && 93.6 & 92.2 & 79.7 & 84.0 & 75.0\\
    PSCCNet && 100K && \textbf{98.2} & 84.7 & 82.9 & 85.5 & 80.6 \\
    Ours && 62K && 95.5 & \textbf{92.8} & \textbf{84.3} & \textbf{87.2} & \textbf{82.1} \\
    \bottomrule
  \end{tabular*}
  \vspace{-0.1in}
  \caption{Comparisons of manipulation localization AUC (\%) scores of different pre-trained models.}
  \label{tab:pretrained}
\end{minipage}
\hfill
\begin{minipage}[t]{0.4\linewidth}
  \vspace{0pt}
  \centering
    \setlength{\tabcolsep}{0pt} 
    \begin{tabular*}{\linewidth}{@{\extracolsep{\fill}}lcccccccc@{}}
    \toprule
    \multirow{2}{*}{\textbf{Method}} & \multicolumn{2}{c}{\textbf{Coverage}} && \multicolumn{2}{c}{\textbf{CASIA}} && \multicolumn{2}{c}{\textbf{NIST16}}  \\
    ~ &  \textbf{AUC} & \textbf{F1} && \textbf{AUC} & \textbf{F1} && \textbf{AUC} & \textbf{F1} \\
    \cmidrule(lr){1-1} \cmidrule{2-9} 
    J-LSTM  & 61.4 & - && - & - && 76.4 & - \\
    H-LSTM  & 71.2 & - && - & - && 79.4 & - \\
    RGB-N  & 81.7 & 43.7 && 79.5 & 40.8 && 93.7 & 72.2 \\
    SPAN  & 93.7 & 55.8 && 83.8 & 38.2 && 96.1 & 58.2 \\
    PSCCNet & 94.1 & 72.3 && 87.5 & 55.4 && 99.6 & 81.9 \\
    Ours & \textbf{95.7} & \textbf{75.8} && \textbf{88.2} & \textbf{57.9} && \textbf{99.6} & \textbf{82.4} \\
    \bottomrule
\end{tabular*}
\vspace{-0.1in}
\caption{Comparison of manipulation localization results using fine-tuned models.}
  \label{tab:finetuned}
\end{minipage} 
\end{table*}

\vspace{0.05in}
\noindent \textbf{Baseline Models} We compare our  method with various baseline models as described below:
\begin{itemize}[leftmargin=*]
\setlength{\itemsep}{5pt}
\item \textbf{J-LSTM}~\cite{bappy2017exploiting}, which employs a hybrid CNN-LSTM architecture to capture the discriminative features between manipulated and non-manipulated regions within a tampered image.
\item \textbf{H-LSTM}~\cite{bappy2019hybrid}, which segments the resampling features extracted by a CNN encoder into patches and adopts an LSTM network to model the transition between different patches for tampering localization.
\item \textbf{RGB-N}~\cite{zhou2018learning}, which adopts an RGB stream and a noise stream in parallel to separately discover tampering features and noise inconsistency within an image.
\item \textbf{ManTraNet}~\cite{wu2019mantra}, which uses a feature extractor to capture the manipulation traces and a local anomaly detection network to localize the manipulated regions.
\item \textbf{SPAN}~\cite{hu2020span}, which leverages a pyramid architecture and models the dependency of image patches through self-attention blocks.
\item \textbf{PSCCNet}~\cite{liu2021pscc}, which employs features at different scales progressively for image tampering localization in a coarse-to-fine manner.
\end{itemize}

\subsection{Image Manipulation Localization}
\label{sec:localization}
Compared with binary tampering detection task, manipulation localization is more challenging because it requires the models to capture more refined forgery features. Following SPAN~\cite{hu2020span} and PSCCNet~\cite{liu2021pscc}, we compare our model with other state-of-the-art tampering localization methods under two settings: 1). training on the synthetic dataset and evaluating on the full test datasets. 2) fine-tuning the pre-trained model on the training split of test datasets and evaluating on their test split.

\vspace{0.1in}

\noindent \textbf{Pre-trained Model} For pre-trained model evaluation, we compare \system with MantraNet~\cite{wu2019mantra}, SPAN~\cite{hu2020span}, and PSCCNet~\cite{liu2021pscc}. We report the AUC scores (\%) in Table~\ref{tab:pretrained}, from which we can observe \system achieves the best localization performance on most datasets. Especially, \system achieves 82.1\% on the real-world dataset IMD20, and outperforms PSCCNet by 1.9\%. This suggests our method owns the superior ability to capture tampering features, and can generalize well to high-quality manipulated images datasets. On Columbia dataset, we surpass SPAN and MaTraNet by 2.0\% and 15.9\%, but are 2.7\% behind PSCCNet. We argue that the reason might be their synthesized training data closely resemble the distribution of the Columbia dataset. This can be further verified by the results in Table~\ref{tab:finetuned}, which demonstrates \system outperforms PSCCNet in both AUC and F1 scores if the model is fine-tuned on Columbia dataset. Moreover, it is worth pointing out \system achieves decent results using less pre-training data compared with other methods.

\vspace{0.1in}

\noindent \textbf{Fine-tuned Model} To compensate for the difference in visual quality between the synthesized datasets and standard datasets, we further fine-tune the pre-trained models on specific datasets, and compare with other methods in Table~\ref{tab:finetuned}. We can observe significant performance gains, which illustrates that \system could capture subtle tampering artifacts through the object-level and patch-level consistency modeling and the multimodal design.

\subsection{Image Manipulation Detection}
\label{sec:detection}
Since most previous studies do not consider tampering detection task, we evaluate our model on CASIA-D introduced by~\cite{liu2021pscc}. Table~\ref{tab:detection} shows the AUC and F1 scores (\%) for detecting manipulated images. The results demonstrate that our model achieves state-of-the-art performance, \ie, 99.70\% in terms of AUC and 97.34\% in F1, which demonstrates the effectiveness of our method to capture manipulation artifacts.

\begin{table}[!ht]
\centering

  \setlength{\tabcolsep}{0pt} 
  \begin{tabular*}{0.7\linewidth}{@{\extracolsep{\fill}}lccc@{}} 
    \toprule
    \textbf{Method}  && \textbf{AUC}  & \textbf{F1} \\
    \cmidrule{1-1} \cmidrule{3-4}
    MantraNet && 59.94 & 56.69 \\
    SPAN && 67.33 & 63.48 \\
    PSCCNet && 99.65 & 97.12 \\
    Ours && \textbf{99.70} & \textbf{97.34} \\
    \bottomrule
\end{tabular*}
\vspace{-0.1in}
\caption{AUC and F1 scores (\%) of tampering detection results on CASIA-D dataset~\cite{liu2021pscc}. The best results are marked in bold.}
 \label{tab:detection}
\end{table}

\subsection{Robustness Evaluation}
\label{sec:robust} 
Following PSCCNet~\cite{liu2021pscc}, we apply different image distortion methods on raw images from NIST16 dataset and evaluate the robustness of our \system. The distortion types include: 1) image scaling with different scales, 2) Gaussian blurring with a kernel size $k$, 3) Gaussian noise with a standard deviation $\sigma$, and 4) JPEG compression with a quality factor $q$. We compare the manipulation localization performance (AUC scores) of our pre-trained models with SPAN~\cite{hu2020span} and PSCCNet~\cite{liu2021pscc} on these corrupted data, and report the results in Table~\ref{tab:robust}. \system demonstrates better robustness against various distortion techniques, especially on compressed images (1.1\% higher than PSCCNet when the quality factor is 100, and 1.0\% higher when the quality factor is 50).

\begin{table}[!ht]
  \centering
    \setlength{\tabcolsep}{0pt} 
    \begin{tabular*}{0.95\linewidth}{@{\extracolsep{\fill}}lcccc@{}}
    \toprule
    \textbf{Distortion} && \textbf{SPAN} & \textbf{PSCCNet} & \textbf{Ours} \\
    \cmidrule{1-1} \cmidrule{3-5} 
    no distortion  && 83.95 & 85.47 & \textbf{87.18}\\
    \cmidrule{1-1} \cmidrule{3-5} 
    Resize (0.78$\times$)  && 83.24 & 85.29 & \textbf{87.17} \Drop{0.01} \\
    Resize (0.25$\times$) &&  80.32 & 85.01 & \textbf{86.33}\Drop{0.85} \\
    \cmidrule{1-1} \cmidrule{3-5} 
    GaussianBlur ($k$=3) && 83.10 & 85.38 & \textbf{85.97} \Drop{1.21} \\
    GaussianBlur ($k$=15) &&  79.15 & 79.93 & \textbf{80.26} \Drop{6.92} \\
    \cmidrule{1-1} \cmidrule{3-5} 
    GaussianNoise ($\sigma$=3) && 75.17 & 78.42 & \textbf{79.58} \Drop{7.60} \\
    GaussianNoise ($\sigma$=15) &&  67.28 & 76.65 & \textbf{78.15} \Drop{9.03} \\
    \cmidrule{1-1} \cmidrule{3-5} 
    JPEGCompress ($q$=100) && 83.59 & 85.40 & \textbf{86.37}\Drop{0.81} \\
    JPEGCompress ($q$=50) &&  80.68 & 85.37 & \textbf{86.24} \Drop{0.94} \\
    \bottomrule
\end{tabular*}
\vspace{-0.1in}
\caption{Localization performance on NIST16 dataset under various distortions. AUC scores are reported (in \%).}
\label{tab:robust}
\end{table}

\begin{figure*}[t]
\begin{minipage}[t]{0.53\textwidth}
  \vspace{0pt}
  \centering
    \renewcommand{\arraystretch}{1.3}
    \setlength{\tabcolsep}{0pt} 
    \begin{tabular*}{\linewidth}{@{\extracolsep{\fill}}lcccccc@{}}
    \toprule
    \multirow{2}{*}{\textbf{Variants}} && \multicolumn{2}{c}{\textbf{CASIA}} && \multicolumn{2}{c}{\textbf{NIST16}} \\
    ~ && \textbf{AUC} & \textbf{F1} && \textbf{AUC} & \textbf{F1} \\
     \cmidrule{1-1} \cmidrule{3-4} \cmidrule{6-7}
     Baseline (EfficientNet-b4) && 70.3 & 37.2 && 77.4 & 51.9 \\
     w/o \textit{HFE} && 75.3 & 51.5 && 88.6 & 63.9 \\
     w/o \textit{BCIM} && 82.7 & 40.1 && 97.2 & 74.5 \\
     vanilla self-attention && 84.8 & 47.6 && 94.5 & 72.1 \\
     Ours && \textbf{88.2} & \textbf{57.9} && \textbf{99.6} & \textbf{82.4} \\
  \bottomrule
  \end{tabular*}
\end{minipage}
\quad\quad\quad
\begin{minipage}[t]{0.4\textwidth}
  \vspace{0pt}
  \centering
  \includegraphics[width=1.0\linewidth]{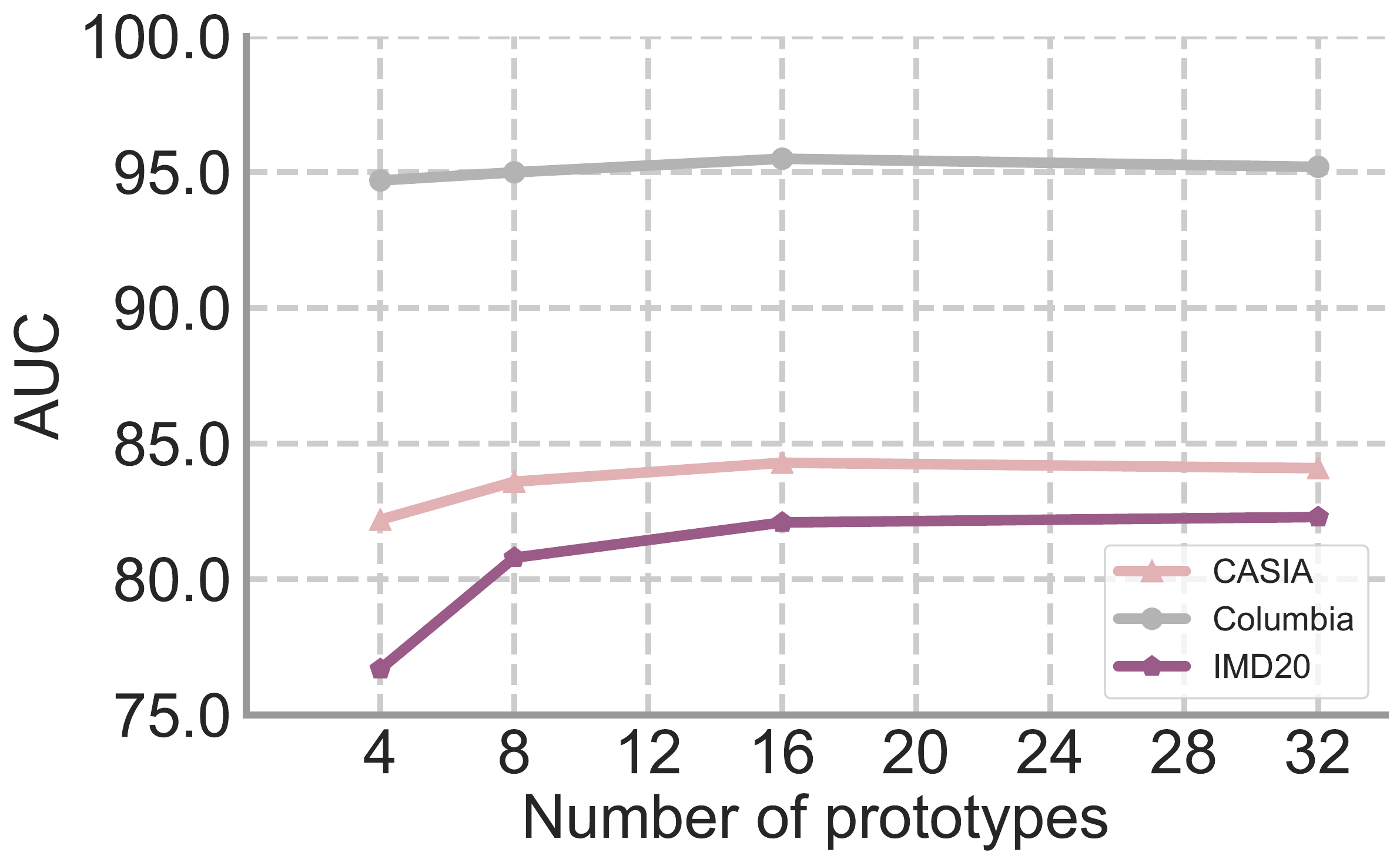}
\end{minipage}
\\
\begin{minipage}[b]{0.53\linewidth}
  \tabcaption{Ablation results on CASIA and NIST16 dataset using different variants of \system. Both AUC and F1 scores (\%) are reported. }
  \label{tab:ablation}
\end{minipage}
\quad\quad\quad
\begin{minipage}[b]{0.4\linewidth}
  \figcaption{AUC scores (\%) of \system with different numbers of object prototypes.} 
  \label{fig:prototype}
\end{minipage}
\end{figure*}

\begin{figure*}[!ht]
  \centering
  \includegraphics[width=1.0\linewidth]{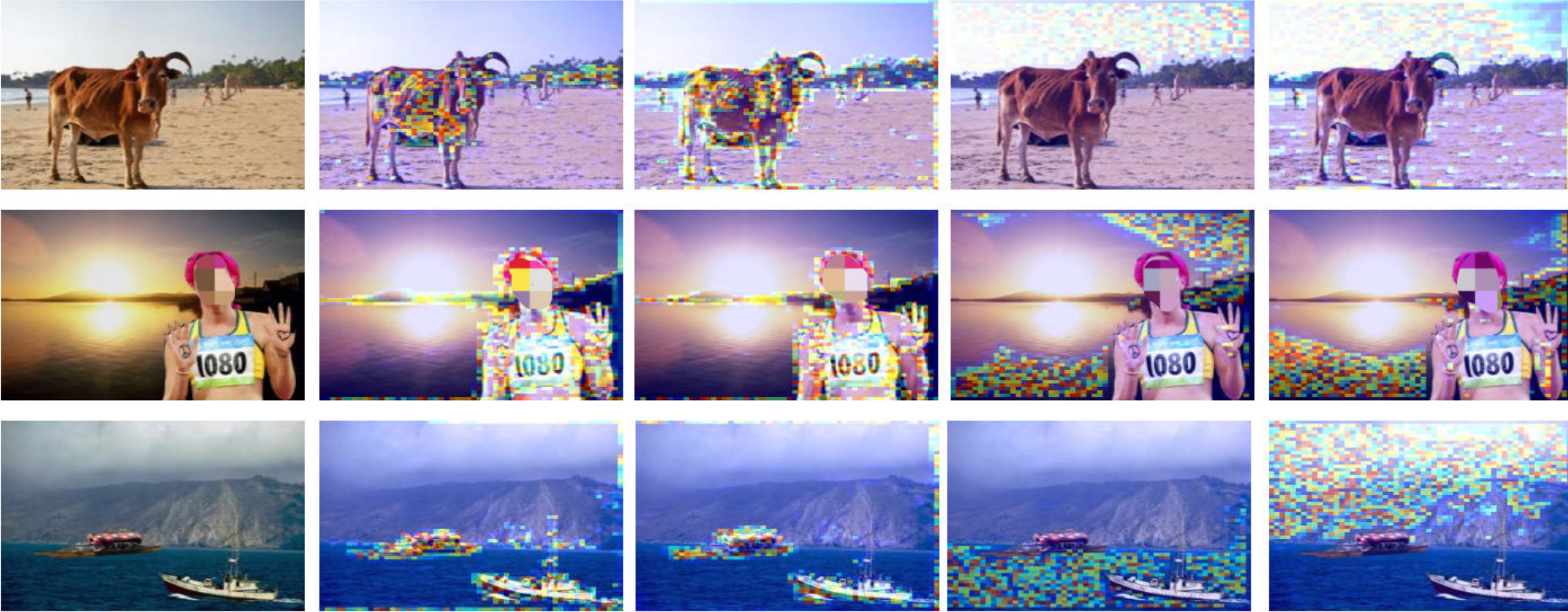}
  \vspace{-0.2in}
  \caption{Visualization of the affinity matrix $A_{i}$ in the first object encoder. From left to right, we display the forged images, two prototypes corresponding to two foreground objects, and two prototypes related to background objects.}
  \label{fig:object}
\end{figure*}

\subsection{Ablation Analysis}
\label{sec:ablation}
The High-frequency Feature Extraction (\textit{HFE}) module of our method is designed to extract the abnormal forgery features in the frequency domain, while the Boundary-sensitivity Contextual Incoherence Modeling (\textit{BCIM}) module is utilized to improve the sharpness of the predicted tampering masks. To evaluate the effectiveness of \textit{HFE} and \textit{BCIM}, we remove them separately from \system and evaluate the tampering localization performance on CASIA and NIST16 dataset. 

The quantitative results are listed in Table~\ref{tab:ablation}. We can observe that without \textit{HFE}, the AUC scores decrease by 14.6\% on CASIA and 11.0\% on NIST16, while without \textit{BCIM}, the AUC scores decrease by 6.2\% on CASIA and 2.4\% on NIST16. The performance degradation validates that the use of \textit{HFE} and \textit{BCIM} effectively improves the performance of our model. Moreover, to illustrate the effectiveness of representations learned by \system, we discard the object representations and replace the stacked object encoders and image decoders with vanilla self-attention blocks. We can observe a significant performance degradation in the third row of Table~\ref{tab:ablation}, \ie, 5\% in terms of AUC and 12.5\% in terms of F1 on NIST16 dataset.

The object prototypes are deployed to represent the visual elements that may appear in an image, which facilitates \system to learn the mid-level semantic features for object-level consistencies modeling. We further conduct experiments to investigate the effect of the number of prototypes ($N$) on model performance. As shown in Figure~\ref{fig:prototype}, there is an overall incremental trend in the tampering location performance as the number of prototypes increases, and the best are achieved on Columbia and CASIA dataset when $N$ is set to 16. 

\begin{figure*}[t]
  \centering
  \includegraphics[width=1\linewidth]{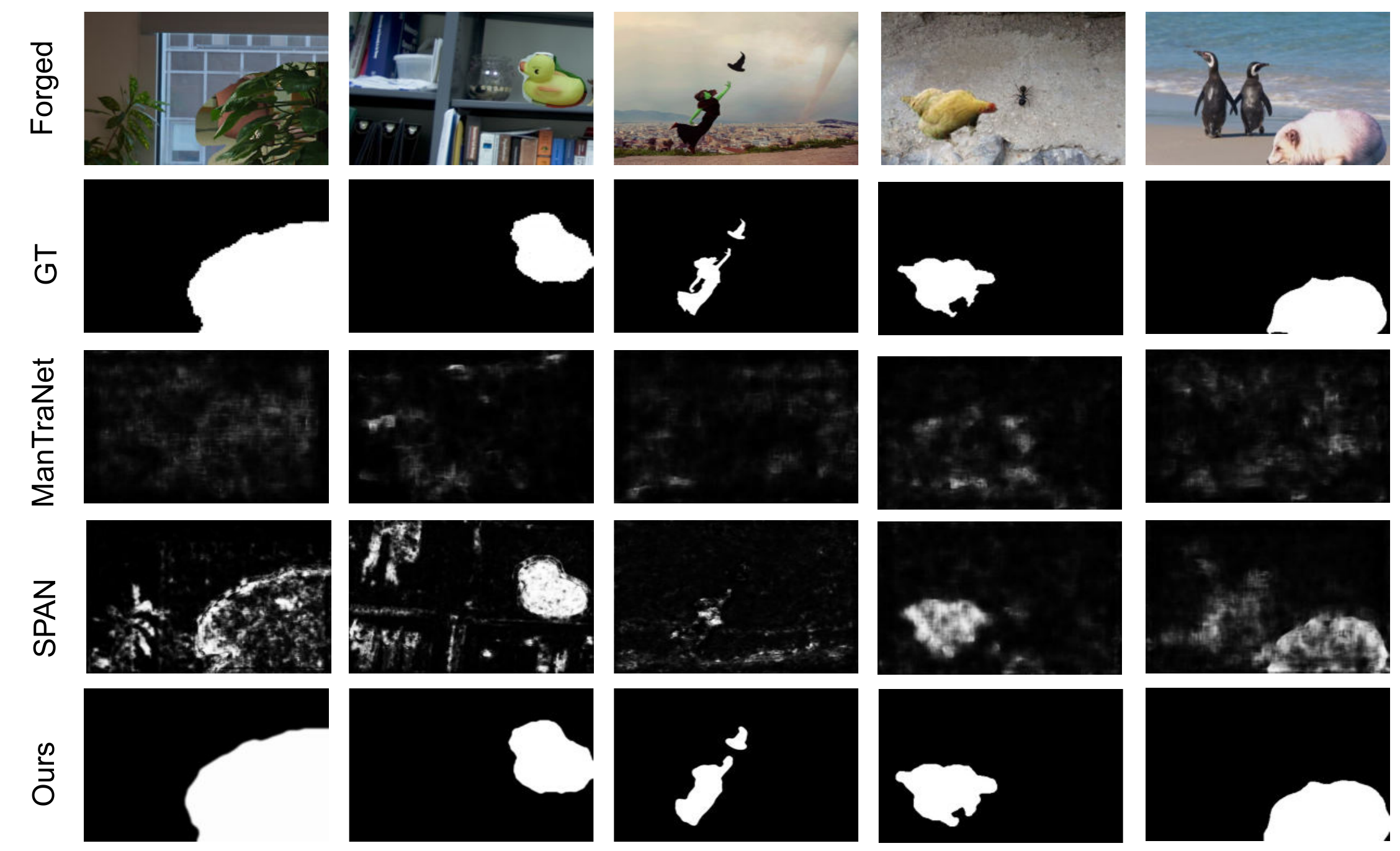}
  \vspace{-0.1in}
  \caption{Visualization of the predicted manipulation mask by different methods. From top to bottom, we show forged images, GT masks, predictions of ManTraNet, SPAN, and ours.}
  \label{fig:com}
\end{figure*}

\subsection{Visualization Results}
\label{sec:vis}
\noindent \textbf{Visualization of object encoder.} We further investigate the behavior of \system qualitatively. Specifically, we average all heads of the affinity matrix $A_{i}$ (Eqn.~\ref{eq:affinity}) in the first object encoder, and then normalize it to [0, 255]. For each image, we visualize the pristine image (column 1), and regions that are attended to by different object prototypes, \eg, column 2 and 3 are two prototypes correspond to two foreground objects while column 4 and 5 relate to background objects. The results in Figure~\ref{fig:object} suggest that through iterative updates, the object representations correspond to meaningful regions in the images, thus contributing to object consistency modeling.

\begin{figure}[!]
  \centering
  \includegraphics[width=\linewidth]{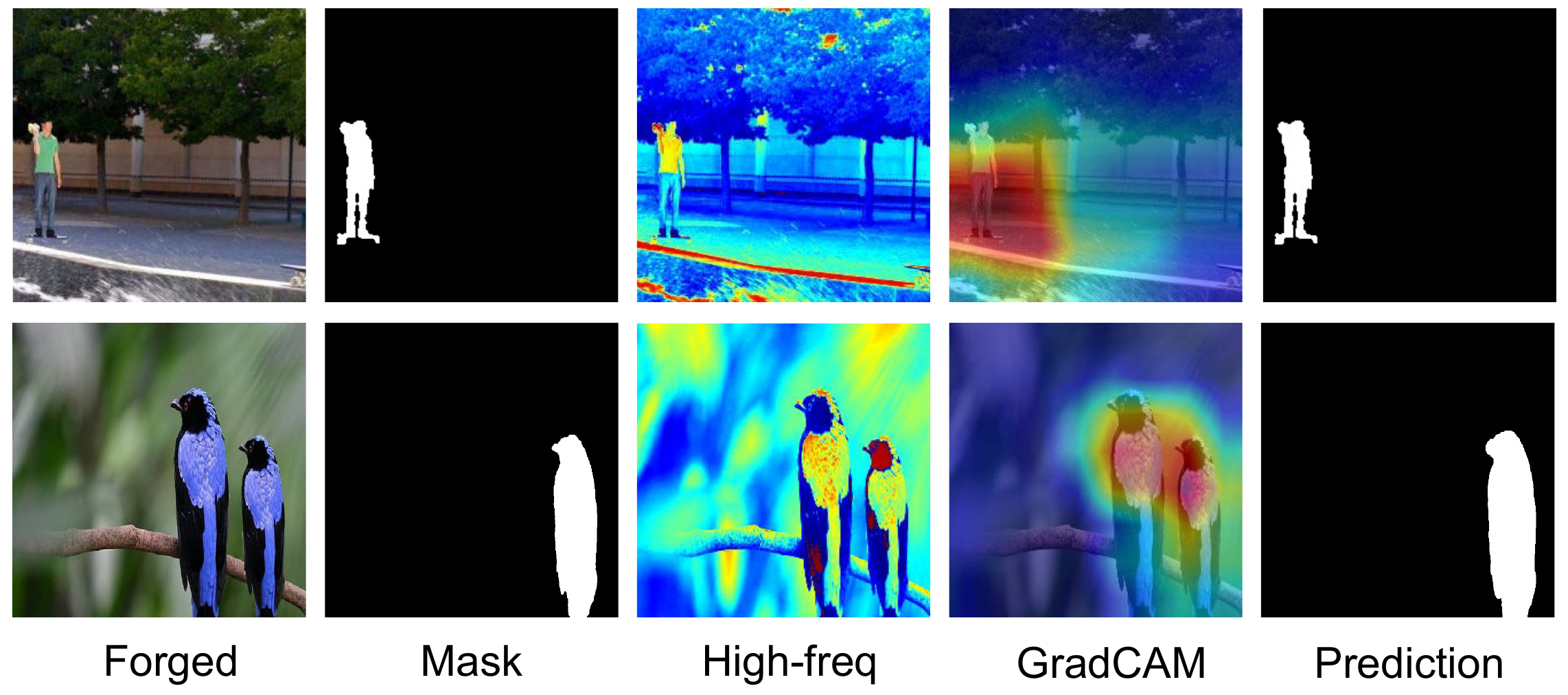}
  \vspace{-0.2in}
  \caption{Visualization of the frequency features. From left to right, we display the forged images, masks, the high-frequency components, GradCAM of the feature maps after conv layer, and \system predictions.}
  \vspace{-0.2in}
  \label{fig:freq}
\end{figure}

\vspace{0.05in}
\noindent \textbf{Qualitative results.} We provide predicted manipulation masks of different methods in Figure~\ref{fig:com}. Since the source code of PSCCNet~\cite{liu2021pscc} is not available, their predictions are not available. The results demonstrate that our method could not only locate the tampering regions more accurately, but also develop more sharp boundaries, which benefits from the inconsistency modeling ability and boundaries sensitivity of \system.

\vspace{0.05in}
\noindent \textbf{Visualization of high-frequency features.} To verify the usefulness of frequency features for tampering detection, we visualize the high-frequency components and \textit{HFE} features using GradCAM~\cite{selvaraju2017grad} (Sec.~\ref{sec:high_freq}) in Figure~\ref{fig:freq}. The results demonstrate that although the forged images are visually natural, the manipulated regions are distinguishable from the untouched areas in frequency domain. 

\subsection{Limitation}
\system faces one potential limitation: when using the pre-trained model to evaluate the performance of tampering localization on Columbia, \system is 2.7\% lower than PSCCNet~\cite{liu2021pscc} in terms of AUC score. The possible reason might be that the pre-training data they use closely resemble the data distribution in the Columbia dataset. Therefore, we believe this problem can be resolved by using more pre-training data.

\section{Conclusion}
\label{sec:conclusion}
We introduced \system, an end-to-end multimodal framework for image tampering detection and localization. To detect subtle manipulation artifacts that are no longer visible in RGB domain, \system extracts forgery features in the frequency domain as complementary information, which are further combined with the RGB features to generate multimodal patch embeddings. Additionally, \system leverages learnable object prototypes as mid-level representations, and alternately updates the object prototypes and patch embeddings with stacked object encoders and patch decoders to model the object-level and patch-level visual consistencies within the images. Extensive experiments on different datasets demonstrate the effectiveness of the proposed method. 

\vspace{0.02in}
\noindent \textbf{Acknowledgement}
This work was supported by National Natural Science Foundation of Project (62072116). Y.-G. Jiang was sponsored in part by ``Shuguang Program'' supported by Shanghai Education Development Foundation and Shanghai Municipal Education Commission (20SG01).

{\small
\bibliographystyle{ieee_fullname}
\bibliography{main}

\begin{thebibliography}{10}\itemsep=-1pt

\bibitem{bai2021visual}
Song Bai, Philip Torr, et~al.
\newblock Visual parser: Representing part-whole hierarchies with transformers.
\newblock {\em arXiv preprint arXiv:2107.05790}, 2021.

\bibitem{bappy2017exploiting}
Jawadul~H Bappy, Amit~K Roy-Chowdhury, Jason Bunk, Lakshmanan Nataraj, and BS
  Manjunath.
\newblock Exploiting spatial structure for localizing manipulated image
  regions.
\newblock In {\em ICCV}, 2017.

\bibitem{bappy2019hybrid}
Jawadul~H Bappy, Cody Simons, Lakshmanan Nataraj, BS Manjunath, and Amit~K
  Roy-Chowdhury.
\newblock Hybrid lstm and encoder--decoder architecture for detection of image
  forgeries.
\newblock {\em TIP}, 2019.

\bibitem{bertasius2021space}
Gedas Bertasius, Heng Wang, and Lorenzo Torresani.
\newblock Is space-time attention all you need for video understanding?
\newblock In {\em ICML}, 2021.

\bibitem{carion2020end}
Nicolas Carion, Francisco Massa, Gabriel Synnaeve, Nicolas Usunier, Alexander
  Kirillov, and Sergey Zagoruyko.
\newblock End-to-end object detection with transformers.
\newblock In {\em ECCV}, 2020.

\bibitem{chen2021local}
Shen Chen, Taiping Yao, Yang Chen, Shouhong Ding, Jilin Li, and Rongrong Ji.
\newblock Local relation learning for face forgery detection.
\newblock In {\em AAAI}, 2021.

\bibitem{cozzolino2015efficient}
Davide Cozzolino, Giovanni Poggi, and Luisa Verdoliva.
\newblock Efficient dense-field copy--move forgery detection.
\newblock {\em TIFS}, 2015.

\bibitem{cozzolino2015splicebuster}
Davide Cozzolino, Giovanni Poggi, and Luisa Verdoliva.
\newblock Splicebuster: A new blind image splicing detector.
\newblock In {\em WIFS}, 2015.

\bibitem{deng2009imagenet}
Jia Deng, Wei Dong, Richard Socher, Li-Jia Li, Kai Li, and Li Fei-Fei.
\newblock Imagenet: A large-scale hierarchical image database.
\newblock In {\em CVPR}, 2009.

\bibitem{dhamo2020semantic}
Helisa Dhamo, Azade Farshad, Iro Laina, Nassir Navab, Gregory~D Hager, Federico
  Tombari, and Christian Rupprecht.
\newblock Semantic image manipulation using scene graphs.
\newblock In {\em CVPR}, 2020.

\bibitem{dong2013casia}
Jing Dong, Wei Wang, and Tieniu Tan.
\newblock Casia image tampering detection evaluation database.
\newblock In {\em ChinaSIP}, 2013.

\bibitem{dosovitskiy2020image}
Alexey Dosovitskiy, Lucas Beyer, Alexander Kolesnikov, Dirk Weissenborn,
  Xiaohua Zhai, Thomas Unterthiner, Mostafa Dehghani, Matthias Minderer, Georg
  Heigold, Sylvain Gelly, Jakob Uszkoreit, and Neil Houlsby.
\newblock An image is worth 16x16 words: Transformers for image recognition at
  scale.
\newblock In {\em ICLR}, 2021.

\bibitem{goodfellow2014generative}
Ian Goodfellow, Jean Pouget-Abadie, Mehdi Mirza, Bing Xu, David Warde-Farley,
  Sherjil Ozair, Aaron Courville, and Yoshua Bengio.
\newblock Generative adversarial nets.
\newblock In {\em NIPS}, 2014.

\bibitem{hendrycks2016gaussian}
Dan Hendrycks and Kevin Gimpel.
\newblock Gaussian error linear units (gelus).
\newblock {\em arXiv preprint arXiv:1606.08415}, 2016.

\bibitem{heo2021rethinking}
Byeongho Heo, Sangdoo Yun, Dongyoon Han, Sanghyuk Chun, Junsuk Choe, and
  Seong~Joon Oh.
\newblock Rethinking spatial dimensions of vision transformers.
\newblock In {\em ICCV}, 2021.

\bibitem{hu2020span}
Xuefeng Hu, Zhihan Zhang, Zhenye Jiang, Syomantak Chaudhuri, Zhenheng Yang, and
  Ram Nevatia.
\newblock Span: Spatial pyramid attention network for image manipulation
  localization.
\newblock In {\em ECCV}, 2020.

\bibitem{huh2018fighting}
Minyoung Huh, Andrew Liu, Andrew Owens, and Alexei~A Efros.
\newblock Fighting fake news: Image splice detection via learned
  self-consistency.
\newblock In {\em ECCV}, 2018.

\bibitem{kingma2013auto}
Diederik~P Kingma and Max Welling.
\newblock Auto-encoding variational bayes.
\newblock {\em arXiv preprint arXiv:1312.6114}, 2013.

\bibitem{kniaz2019point}
Vladimir~V Kniaz, Vladimir Knyaz, and Fabio Remondino.
\newblock The point where reality meets fantasy: Mixed adversarial generators
  for image splice detection.
\newblock 2019.

\bibitem{li2020manigan}
Bowen Li, Xiaojuan Qi, Thomas Lukasiewicz, and Philip~HS Torr.
\newblock Manigan: Text-guided image manipulation.
\newblock In {\em CVPR}, 2020.

\bibitem{lin2014microsoft}
Tsung-Yi Lin, Michael Maire, Serge Belongie, James Hays, Pietro Perona, Deva
  Ramanan, Piotr Doll{\'a}r, and C~Lawrence Zitnick.
\newblock Microsoft coco: Common objects in context.
\newblock In {\em ECCV}, 2014.

\bibitem{liu2021pscc}
Xiaohong Liu, Yaojie Liu, Jun Chen, and Xiaoming Liu.
\newblock Pscc-net: Progressive spatio-channel correlation network for image
  manipulation detection and localization.
\newblock {\em arXiv preprint arXiv:2103.10596}, 2021.

\bibitem{liu2021video}
Ze Liu, Jia Ning, Yue Cao, Yixuan Wei, Zheng Zhang, Stephen Lin, and Han Hu.
\newblock Video swin transformer.
\newblock {\em arXiv preprint arXiv:2106.13230}, 2021.

\bibitem{meng2021adavit}
Lingchen Meng, Hengduo Li, Bor-Chun Chen, Shiyi Lan, Zuxuan Wu, Yu-Gang Jiang,
  and Ser-Nam Lim.
\newblock Adavit: Adaptive vision transformers for efficient image recognition.
\newblock In {\em CVPR}, 2022.

\bibitem{mirza2014conditional}
Mehdi Mirza and Simon Osindero.
\newblock Conditional generative adversarial nets.
\newblock {\em arXiv preprint arXiv:1411.1784}, 2014.

\bibitem{nazeri2019edgeconnect}
Kamyar Nazeri, Eric Ng, Tony Joseph, Faisal Qureshi, and Mehran Ebrahimi.
\newblock Edgeconnect: Structure guided image inpainting using edge prediction.
\newblock In {\em ICCVW}, 2019.

\bibitem{nimble16}
Nist.
\newblock Nimble 2016 datasets.
\newblock
  \url{https://www.nist.gov/itl/iad/mig/nimble-challenge-2017-evaluation},
  2016.

\bibitem{novozamsky2020imd2020}
Adam Novozamsky, Babak Mahdian, and Stanislav Saic.
\newblock Imd2020: A large-scale annotated dataset tailored for detecting
  manipulated images.
\newblock In {\em WACVW}, 2020.

\bibitem{park2020swapping}
Taesung Park, Jun-Yan Zhu, Oliver Wang, Jingwan Lu, Eli Shechtman, Alexei~A.
  Efros, and Richard Zhang.
\newblock Swapping autoencoder for deep image manipulation.
\newblock In {\em NIPS}, 2020.

\bibitem{pathak2016context}
Deepak Pathak, Philipp Krahenbuhl, Jeff Donahue, Trevor Darrell, and Alexei~A
  Efros.
\newblock Context encoders: Feature learning by inpainting.
\newblock In {\em CVPR}, 2016.

\bibitem{qian2020thinking}
Yuyang Qian, Guojun Yin, Lu Sheng, Zixuan Chen, and Jing Shao.
\newblock Thinking in frequency: Face forgery detection by mining
  frequency-aware clues.
\newblock In {\em ECCV}, 2020.

\bibitem{rao2016deep}
Yuan Rao and Jiangqun Ni.
\newblock A deep learning approach to detection of splicing and copy-move
  forgeries in images.
\newblock In {\em WIFS}, 2016.

\bibitem{razavi2019generating}
Ali Razavi, Aaron van~den Oord, and Oriol Vinyals.
\newblock Generating diverse high-resolution images with vq-vae.
\newblock In {\em ICLR Workshop}, 2019.

\bibitem{selvaraju2017grad}
Ramprasaath~R Selvaraju, Michael Cogswell, Abhishek Das, Ramakrishna Vedantam,
  Devi Parikh, and Dhruv Batra.
\newblock Grad-cam: Visual explanations from deep networks via gradient-based
  localization.
\newblock In {\em ICCV}, 2017.

\bibitem{shi2000normalized}
Jianbo Shi and Jitendra Malik.
\newblock Normalized cuts and image segmentation.
\newblock {\em TPAMI}, 2000.

\bibitem{tan2019efficientnet}
Mingxing Tan and Quoc Le.
\newblock Efficientnet: Rethinking model scaling for convolutional neural
  networks.
\newblock In {\em ICML}, 2019.

\bibitem{vaswani2017attention}
Ashish Vaswani, Noam Shazeer, Niki Parmar, Jakob Uszkoreit, Llion Jones,
  Aidan~N. Gomez, Lukasz Kaiser, and Illia Polosukhin.
\newblock Attention is all you need.
\newblock In {\em NIPS}, 2017.

\bibitem{vinker2020deep}
Yael Vinker, Eliahu Horwitz, Nir Zabari, and Yedid Hoshen.
\newblock Deep single image manipulation.
\newblock {\em arXiv preprint arXiv:2007.01289}, 2020.

\bibitem{wang2021m2tr}
Junke Wang, Zuxuan Wu, Jingjing Chen, and Yu-Gang Jiang.
\newblock M2tr: Multi-modal multi-scale transformers for deepfake detection.
\newblock {\em arXiv preprint arXiv:2104.09770}, 2021.

\bibitem{wang2021efficient}
Junke Wang, Xitong Yang, Hengduo Li, Zuxuan Wu, and Yu-Gang Jiang.
\newblock Efficient video transformers with spatial-temporal token selection.
\newblock {\em arXiv preprint arXiv:2111.11591}, 2021.

\bibitem{wang2021bevt}
Rui Wang, Dongdong Chen, Zuxuan Wu, Yinpeng Chen, Xiyang Dai, Mengchen Liu,
  Yu-Gang Jiang, Luowei Zhou, and Lu Yuan.
\newblock Bevt: Bert pretraining of video transformers.
\newblock In {\em CVPR}, 2022.

\bibitem{wen2016coverage}
Bihan Wen, Ye Zhu, Ramanathan Subramanian, Tian-Tsong Ng, Xuanjing Shen, and
  Stefan Winkler.
\newblock Coverage—a novel database for copy-move forgery detection.
\newblock In {\em ICIP}, 2016.

\bibitem{wu2019mantra}
Yue Wu, Wael AbdAlmageed, and Premkumar Natarajan.
\newblock Mantra-net: Manipulation tracing network for detection and
  localization of image forgeries with anomalous features.
\newblock In {\em CVPR}, 2019.

\bibitem{xiao2021early}
Tete Xiao, Piotr Dollar, Mannat Singh, Eric Mintun, Trevor Darrell, and Ross
  Girshick.
\newblock Early convolutions help transformers see better.
\newblock In {\em NIPS}, 2021.

\bibitem{yuan2021tokens}
Li Yuan, Yunpeng Chen, Tao Wang, Weihao Yu, Yujun Shi, Zi-Hang Jiang,
  Francis~EH Tay, Jiashi Feng, and Shuicheng Yan.
\newblock Tokens-to-token vit: Training vision transformers from scratch on
  imagenet.
\newblock In {\em ICCV}, 2021.

\bibitem{zhou2018learning}
Peng Zhou, Xintong Han, Vlad~I Morariu, and Larry~S Davis.
\newblock Learning rich features for image manipulation detection.
\newblock In {\em CVPR}, 2018.

\bibitem{zhu2017unpaired}
Jun-Yan Zhu, Taesung Park, Phillip Isola, and Alexei~A Efros.
\newblock Unpaired image-to-image translation using cycle-consistent
  adversarial networks.
\newblock In {\em ICCV}, 2017.

\bibitem{zhu2018deep}
Xinshan Zhu, Yongjun Qian, Xianfeng Zhao, Biao Sun, and Ya Sun.
\newblock A deep learning approach to patch-based image inpainting forensics.
\newblock {\em Signal Processing: Image Communication}, 2018.

\bibitem{zhu2020deformable}
Xizhou Zhu, Weijie Su, Lewei Lu, Bin Li, Xiaogang Wang, and Jifeng Dai.
\newblock Deformable {\{}detr{\}}: Deformable transformers for end-to-end
  object detection.
\newblock In {\em ICLR}, 2021.

\end{thebibliography}
}

\end{document}